# Sensitivity analysis for finite Markov chains in discrete time

**Gert de Cooman, Filip Hermans** and **Erik Quaeghebeur**
SYSTeMS Research Group
Ghent University, Belgium
{gert.decooman,filip.hermans,erik.quaeghebeur}@UGent.be

## Abstract

When the initial and transition probabilities of a finite Markov chain in discrete time are not well known, we should perform a sensitivity analysis. This is done by considering as basic uncertainty models the so-called *credal sets* that these probabilities are known or believed to belong to, and by allowing the probabilities to vary over such sets. This leads to the definition of an *imprecise Markov chain*. We show that the time evolution of such a system can be studied very efficiently using so-called *lower* and *upper expectations*. We also study how the inferred credal set about the state at time $n$ evolves as $n \to \infty$: under quite unrestrictive conditions, it converges to a uniquely invariant credal set, regardless of the credal set given for the initial state. This leads to a non-trivial generalisation of the classical Perron–Frobenius Theorem to imprecise Markov chains.

## 1 Setting the stage

One convenient way to model uncertain dynamical systems is to describe them as Markov chains. These have been studied in great detail, and their properties are well known. However, in many practical situations, it remains a challenge to accurately identify the transition probabilities in the Markov chain: the available information about physical systems is often imprecise and uncertain. Describing a real-life dynamical system as a Markov chain will therefore often involve unwarranted precision, and may lead to conclusions not supported by the available information.

For this reason, it seems quite useful to perform probabilistic robustness studies, or sensitivity analyses, for Markov chains. This is especially relevant in decision-making applications. Many researchers in Markov Chain Decision Making (White and Eldeib, 1994; Harmanec, 2002; Nilim and El Ghaoui, 2005; Itoh and Nakamura, 2007)—inspired by Satia and Lave's (1973) original work—have paid attention to this issue of 'imprecision' in Markov chains.

Early work on the more mathematical aspects of modelling such 'imprecision' in Markov chains was done by Kozine and Utkin (2002). Armed with linear programming techniques, these authors also performed an experimental study of the limit behaviour of Markov chains with uncertain transition probabilities. More recently, Škulj (2006, 2007) has begun a formal study of the time evolution and limit behaviour of such systems.

All these approaches use *sets of probabilities* to deal with the imprecision in the transition probabilities. When these probabilities are not well known, they are assumed to belong to certain sets, and robustness analyses are performed by allowing the transition probabilities to vary over such sets. As we shall see, this approach leads to a number of computational difficulties, which we show can be overcome by tackling the same problem from another angle, using lower and upper expectations, rather than sets of probabilities.

In the rest of this Introduction, we give an overview of the theory of classical Markov chains and formulate the classical Perron–Frobenius theorem. Then, in Sections 2 and 3, we introduce imprecise Markov chains and generalise many aspects of the classical theory. In Section 4, we generalize the Perron–Frobenius theorem. We discuss a number of theoretical and numerical examples in Section 5, and we give perspectives for further research in the Conclusions.

### 1.1 Analysis of classical Markov chains

Consider a finite Markov chain in discrete time, where at times $n = 1, 2, 3, \dots, N$, $N \in \mathbb{N}$ the *state* $X(n)$ of a system can assume any value in a finite set $\mathscr{X}$. Here $\mathbb{N}$ denotes the set of non-zero natural numbers, and $N$ is the time horizon. The time evolution of such a system can be modelled as if it traversed a so-called *event tree* (Shafer, 1996). An example of such a tree for $\mathscr{X} = \{a, b\}$ and $N = 3$ is given in Figure 1. The *situations*, or nodes, of the tree have the form $x_{1:k} := (x_1, \dots, x_k) \in \mathscr{X}^k$, $k = 0, 1, \dots, N$. For $k = 0$ there is some abuse of notation as we let $\mathscr{X}^0 := \{\Box\}$, where $\Box$ is the so-called *initial situation*, or root of the tree. In the *cuts* $\mathscr{X}^n$ of $\square$, the value of the state $X(n)$ at time $n$ is revealed.

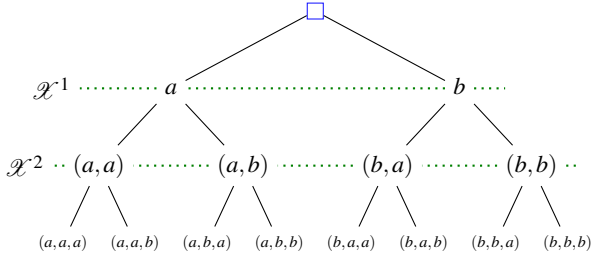

Figure 1: The event tree for the time evolution of system that can be in two states, *a* and *b*, and can change state at time instants $n = 1, 2$. Also depicted are the respective cuts $\mathscr{X}^1$ and $\mathscr{X}^2$ of $\square$ where the states at times 1 and 2 are revealed.

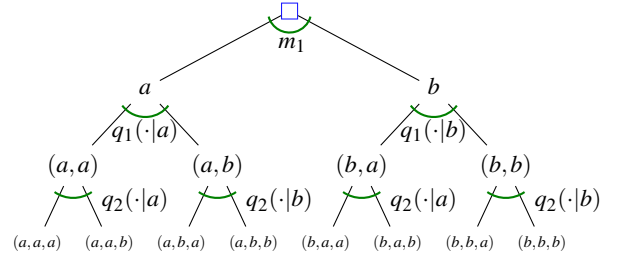

Figure 2: The probability tree for the time evolution of a Markov chain that can be in two states, *a* and *b*, and can change state at each time instant $n = 1, 2$.

In a classical analysis, it is generally assumed that we have: (i) a probability distribution over the initial state $X(1)$, in the form of a probability mass function $m_1$ on $\mathscr{X}$; and (ii) for each situation $x_{1:n}$ that the system can be in at time $n$, a probability distribution over the next state $X(n+1)$, in the form of a probability mass function $q(\cdot|x_{1:n})$ on $\mathscr{X}$. This means that in each non-terminal situation[1] $x_{1:n}$ of the event tree, we have a *local* probability model telling us about the probabilities of each of its child nodes. This turns the event tree into a so-called *probability tree*; see Shafer (1996, Chapter 3) and Kemeny and Snell (1976, Section 1.9).

The probability tree for a Markov chain is special, because the *Markov Condition* states that when the system jumps from state $X(n) = x_n$ to a new state $X(n+1)$, where the system goes to will only depend on the state $X(n) = x_n$ the system was in at time $n$, and not on its states $X(k) = x_k$ at previous times $k = 1, 2, \ldots, n-1$. In other words:

$$q(\cdot|x_{1:n}) = q_n(\cdot|x_n), x_{1:n} \in \mathscr{X}^n, n = 1, \ldots, N-1, \quad (1)$$

where $q_n(\cdot|x_n)$ is some probability mass function on $\mathscr{X}$. The Markov chain may be non-stationary, as the transition probabilities are allowed to depend explicitly on the time $n$. Figure 2 gives an example of a probability tree for a Markov chain with $\mathscr{X} = \{a, b\}$ and $N = 3$.

With the local probability mass functions $m_1$ and $q_n(\cdot|x_n)$ we associate the linear real-valued *expectation functionals* $E_1$ and $E_n(\cdot|x_n)$, given, for all real-valued maps $h$ on $\mathscr{X}$, by

$$E_1(h) := \sum_{x_1 \in \mathscr{X}} h(x_1) m_1(x_1),$$

$$E_n(h|x_n) := \sum_{x_{n+1} \in \mathscr{X}} h(x_{n+1}) q_n(x_{n+1}|x_n).$$

In any probability tree, probabilities and expectations can be calculated very efficiently using backwards recursion. Suppose that in situation $x_{1:n}$, we want to calculate the conditional expectation $E(f|x_{1:n})$ of some real-valued function $f$ on $\mathscr{X}^N$ that may depend on the values of the states $X(1)$ to $X(N)$. Let us indicate briefly how this is done, also taking into account the simplifications due to the Markov Condition (1). A prominent part is played by the so-called *transition operators* $T_n$ and $\mathbb{T}_n$. Consider the linear space $\mathscr{L}(\mathscr{X})$ of all real-valued maps on $\mathscr{X}$. Then the linear operator $T_n \colon \mathscr{L}(\mathscr{X}) \to \mathscr{L}(\mathscr{X})$ is defined by

$$T_n h(x_n) := E_n(h|x_n) = \sum_{x_{n+1} \in \mathscr{X}} h(x_{n+1}) q_n(x_{n+1}|x_n) \quad (2)$$

for all real-valued maps $h$ on $\mathscr{X}$. In other words, $T_n h$ is the real-valued map on $\mathscr{X}$ whose value $T_n h(x_n)$ in $x_n \in \mathscr{X}$ is the conditional expectation of the random variable $h(X(n+1))$, given that the system is in state $x_n$ at time $n$. More generally, we also consider the linear maps $\mathbb{T}_n$ from $\mathscr{L}(\mathscr{X}^{n+1})$ to $\mathscr{L}(\mathscr{X}^n)$, defined by

$$\mathbb{T}_n f(x_{1:n}) := T_n f(x_{1:n}, \cdot)(x_n) = E_n(f(x_{1:n}, \cdot)|x_n)$$
$$= \sum_{x_{n+1} \in \mathscr{X}} f(x_{1:n}, x_{n+1}) q_n(x_{n+1}|x_n) \quad (3)$$

for all $x_{1:n}$ in $\mathscr{X}^n$ and all real-valued maps $f$ on $\mathscr{X}^{n+1}$. We begin with the expectation $E(f|x_{1:n})$ for $n = N-1$:

$$E(f|x_{1:N-1}) = \sum_{x_N \in \mathscr{X}} f(x_{1:N-1}, x_N) q(x_N|x_{1:N-1})$$
$$= \sum_{x_N \in \mathscr{X}} f(x_{1:N-1}, x_N) q_{N-1}(x_N|x_{N-1})$$
$$= \mathbb{T}_{N-1} f(x_{1:N-1}),$$

where the second inequality follows from the Markov Condition (1), and the third from Eq. (3). Similar arguments for $n = N - 2$ and the Law of Iterated Expectations yield:

$$E(f|x_{1:N-2}) = E(E(f(x_{1:N-2}, \cdot, \cdot)|x_{1:N-2}, \cdot)|x_{1:N-2})$$
$$= \mathbb{T}_{N-2} \mathbb{T}_{N-1} f(x_{1:N-2}).$$

---

[1] A *non-terminal* situation is a node of the tree that is not a leaf.

Repeating this argument leads to the backwards recursion formulae (for $n = 1, \ldots, N-1$)

$$E(f|x_{1:n}) = \mathbb{T}_n \mathbb{T}_{n+1} \ldots \mathbb{T}_{N-1} f(x_{1:n}) \quad (4)$$

$$E(f) := E(f|\square) = E_1(\mathbb{T}_1 \mathbb{T}_2 \ldots \mathbb{T}_{N-1} f). \quad (5)$$

In these formulae, $f$ is any real-valued function on $\mathscr{X}^N$. If we let $f$ be the indicator functions $I_{\{x_{1:N}\}}$ of the singletons $\{x_{1:N}\}$, these formulae allow us for instance to calculate the joint probability mass function $p(x_{1:N}) = E(I_{\{x_{1:N}\}})$ for all the variables $X(1)$, ..., $X(N)$. We can also use them to find the conditional mass functions $p(x_{n+1:N}|x_n) = p(x_{n+1:N}|x_{1:n}) := E(I_{\{x_{1:N}\}}|x_{1:n})$.

### 1.2 The Perron–Frobenius Theorem for classical Markov chains

We are especially interested in the case of a *stationary* Markov chain, and in the (marginal) expectation $E_n(h)$ of a real-valued function $h$ (on $\mathscr{X}$) that depends only on the state $X(n)$ at time $n$. Here, Eq. (5) becomes

$$E_n(h) := E_1(\mathrm{T}^{n-1}h), \qquad (6)$$

where $\mathrm{T} := \mathrm{T}_1 = \mathrm{T}_2 = \cdots = \mathrm{T}_{N-1}$, and where we denote by $\mathrm{T}^k$ the $k$-fold composition of $\mathrm{T}$ with itself; in particular, $\mathrm{T}^0$ is the identity operator on $\mathscr{L}(\mathscr{X})$. If we let $h = I_{\{x_n\}}$, this allows us to find the probability mass function $m_n(x_n) = E_n(I_{\{x_n\}})$ for the state $X(n)$. Under some restrictions on the transition operator $\mathrm{T}$, the classical Perron–Frobenius Theorem then tells us that, as $n$ and the time horizon $N$ recede to infinity, this probability mass function converges to some limit, independently of the initial probability mass function $m_1$; see Kemeny and Snell (1976, Theorem 4.1.6) and Luenberger (1979, Chapter 6). In terms of expectation functionals and transition operators:

**Theorem 1** (Classical Perron–Frobenius Theorem, Expectation Form). *Consider a stationary Markov chain with finite state set $\mathscr{X}$ and transition operator $\mathrm{T}$. Suppose that $\mathrm{T}$ is regular, meaning that there is some $k > 0$ such that $\min \mathrm{T}^k I_{\{x_k\}} > 0$ for all $x_k$ in $\mathscr{X}$. Then for every initial expectation operator $E_1$, the expectation operator $E_n = E_1 \circ \mathrm{T}^{n-1}$ for the state at time n converges point-wise to the same limit expectation operator $E_\infty$: for all $h \in \mathscr{L}(\mathscr{X})$,*

$$\lim_{n \to \infty} E_n(h) = \lim_{n \to \infty} E_1(\mathrm{T}^{n-1}h) = E_\infty(h).$$

*Moreover, the limit expectation $E_\infty$ is the only $\mathrm{T}$-invariant expectation on $\mathscr{L}(\mathscr{X})$, in the sense that $E_\infty = E_\infty \circ \mathrm{T}$.*

## 2 Towards imprecise Markov chains

The treatment above rests on the assumption that the initial probabilities and the transition probabilities are precisely known. If such is not the case, then it seems necessary to perform some kind of sensitivity analysis, in order to find out to what extent any conclusions we might reach using such a treatment, depend on the actual values of these probabilities.

A very general way of performing a sensitivity analysis for probabilities involves calculations with closed convex sets of probability mass functions, also called *credal sets*, rather than with single probability measures. Let $\Sigma_\mathscr{X}$ denote the set of all probability mass functions on $\mathscr{X}$, an $(|\mathscr{X}|-1)$-dimensional unit simplex in the $|\mathscr{X}|$-dimensional linear space $\mathbb{R}^\mathscr{X}$, then $\{m \in \Sigma_\mathscr{X} : (\forall x \in \mathscr{X})(m(x) \leq \frac{1}{2})\}$ is a credal set, but $\{m \in \Sigma_\mathscr{X} : (\exists x \in \mathscr{X})(m(x) \geq \frac{1}{2})\}$ is not.

There is a growing body of literature on this interesting and fairly new area of *imprecise probabilities*, starting with the publication of Walley's (1991) seminal work. We refer to the literature (Walley, 1991, 1996; Weichselberger, 2001; De Cooman and Miranda, 2007) for more details and discussion.

Specifying a closed convex set $\mathscr{P}$ of probability mass functions $p$ on a finite set $\mathscr{Y}$ is equivalent (Walley, 1991, Section 3.4.1) to specifying its *lower* and *upper expectation* (functionals) $\underline{E}_\mathscr{P} \colon \mathscr{L}(\mathscr{Y}) \to \mathbb{R}$ and $\overline{E}_\mathscr{P} \colon \mathscr{L}(\mathscr{Y}) \to \mathbb{R}$, defined by

$$\underline{E}_\mathscr{P}(g) := \min\{E_p(g) \colon p \in \mathscr{P}\}$$
$$\overline{E}_\mathscr{P}(g) := \max\{E_p(g) \colon p \in \mathscr{P}\}$$

for real-valued maps $g$ on $\mathscr{Y}$, where $E_p(g) = \sum_{y \in \mathscr{Y}} g(y)p(y)$ is the expectation of $g$ associated with the probability mass function $p$. In a sensitivity analysis, such functionals are quite useful, because they give tight lower and upper bounds on the expectation of any real-valued map. Since the functionals $\underline{E}_\mathscr{P}$ and $\overline{E}_\mathscr{P}$ are *conjugate* in the sense that $\underline{E}_\mathscr{P}(g) = -\overline{E}_\mathscr{P}(-g)$ for all real-valued maps $g$ on $\mathscr{Y}$, one is completely determined if the other is known. Below, we concentrate on upper expectations.

What is the upshot of all this for the Markov chain problem we are considering here? First of all, in the initial situation $\square$, corresponding to time $n = 0$, rather than a single initial probability mass function $m_1$, we now have a local credal set $\mathscr{M}_1$ of candidate mass functions $m_1$ for the state $X(1)$ that the system will be in at time $k = 1$. We denote by $\overline{E}_1$ the upper expectation associated with $\mathscr{M}_1$:

$$\overline{E}_1(h) := \max\left\{\sum_{x \in \mathscr{X}} h(x)m_1(x) \colon m_1 \in \mathscr{M}_1\right\}$$

for all $h \in \mathscr{L}(\mathscr{X})$. Also, in any situation $x_{1:n} \in \mathscr{X}^n$, corresponding to time $n = 1, 2, \ldots, N-1$, instead of a single transition mass function $q_n(\cdot|x_n)$, we now have a local credal set $\mathscr{Q}_n(\cdot|x_n)$ of candidate conditional mass functions $q_n(\cdot|x_n)$ for the state $X(n+1)$ that the system will be in at time $n+1$. We denote by $\overline{E}_n(\cdot|x_n)$ the upper expectation associated with $\mathscr{Q}_n(\cdot|x_n)$, i.e., for all $h \in \mathscr{L}(\mathscr{X})$:

$$\overline{E}_n(h|x_n) := \max\left\{\sum_{x \in \mathscr{X}} h(x)q(x) \colon q \in \mathscr{Q}_n(\cdot|x_n)\right\}. \qquad (7)$$

We call the resulting model an *imprecise Markov chain*. Figure 3 gives an example of an imprecise Markov chain probability tree. A classical, or *precise*, Markov chain is an imprecise one with credal sets that are singletons.

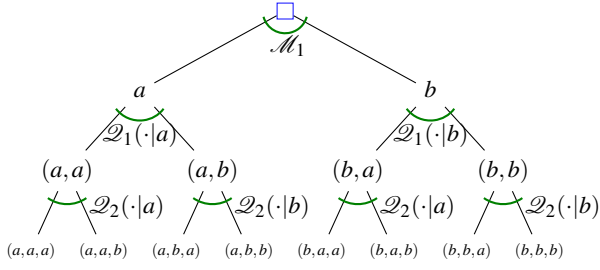

Figure 3: The tree for the time evolution of an imprecise Markov chain that can be in two states, $a$ and $b$, and can change state at each time instant $n = 1, 2$.

How, then, is a sensitivity analysis typically performed (Kozine and Utkin, 2002; Škulj, 2006, 2007) for such an imprecise Markov chain? We choose, in each non-terminal situation $x_{1:k}$ of the above-mentioned event tree, a local transition probability mass $q(\cdot|x_{1:k})$ in the set of possible candidates $\mathcal{Q}_k(\cdot|x_k)$.[2] For $k = 0$, we get the initial situation $\square$, where we choose some element $m_1$ in the set of possible candidates $\mathcal{M}_1$. We then obtain a so-called *compatible probability tree*, for which we may calculate all (conditional) expectations and probability mass functions:

$$E(f|x_{1:n}) = \sum_{x_{n+1:N} \in \mathcal{X}^{N-n}} f(x_{1:n}, x_{n+1:N}) \prod_{k=n}^{N-1} q(x_{k+1}|x_{1:k}), \quad (8)$$

$$E(f) = \sum_{x_{1:N} \in \mathcal{X}^N} f(x_{1:N}) m_1(x_1) \prod_{k=1}^{N-1} q(x_{k+1}|x_{1:k}), \quad (9)$$

for $n = 1, \ldots, N-1$, and for all real-valued maps $f$ on $\mathcal{X}^N$. As we have just come to realise, the probability trees that are compatible with an imprecise Markov chain are no longer necessarily (precise) Markov chains themselves. It is still possible to calculate the $E(f|x_{1:n})$ and $E(f)$ in Eqs. (8) and (9) using backwards recursion (Shafer, 1996, Chapter 3), but the formulae for doing so will be more complicated than the ones for precise Markov chains given by Eqs. (4) and (5).

If we repeat this for every other choice of the $m_1$ in $\mathcal{M}_1$ and the $q(\cdot|x_{1:k})$ in $\mathcal{Q}_k(\cdot|x_k)$, we end up with an infinity of compatible probability trees, for which the associated (conditional) expectations and probability mass functions turn out to be closed convex sets. We denote their corresponding upper expectation functionals on $\mathcal{L}(\mathcal{X}^N)$ by $\overline{E}(\cdot|x_{1:n})$ and $\overline{E}$. These upper expectations, and the conjugate lower expectations, are the final aim of our sensitivity analysis.

The procedure we have just described is computationally very complex. When the closed convex sets $\mathcal{M}_1$ and $\mathcal{Q}_k(\cdot|x)$ each have a finite number of extreme points (are polytopes), we can limit ourselves to working with these sets of extreme points, rather than with the infinite sets themselves. But even then, the computational complexity of this approach will generally be exponential in the number of time steps.

---

[2]These local transition probability masses themselves depend on the situation $x_{1:k}$ they are attached to, but the sets $\mathcal{Q}_k(\cdot|x_k)$ they are chosen from only depend on the last state $x_k$.

However, we prove in Section 3 that the upper expectations $\overline{E}$ and $\overline{E}(\cdot|x_{1:n})$ associated with the closed convex sets of (conditional) probability mass functions for the compatible probability trees of an imprecise Markov chain can be calculated in the same way as the expectations $E$ and $E(\cdot|x_{1:n})$ in a precise one: using counterparts of the backwards recursion formulae (4)–(6). Because of this, making inferences about the mass function of the state at time $n$, i.e., finding the upper envelope $\overline{E}_n$ of the $E_n$ given in Eq. (6) *now has a complexity that is linear, rather than exponential, in the number of time steps $n$.* This is our first contribution.

Our second contribution in this paper is a Perron–Frobenius Theorem for a special class of so-called regular stationary imprecise Markov chains: in Section 4 we prove a generalisation of Theorem 1, which tells us that under the fairly weak condition of regularity, the upper expectation operators $\overline{E}_n$ converge to limits that do not depend on the initial upper expectation operators $\overline{E}_1$.

## 3 Sensitivity analysis of imprecise Markov chains

We are now ready to take our most important step: the backwards recursion formulae for the conditional and joint upper expectations in an imprecise Markov chain. We first define *upper transition operators* $\overline{T}_n$ and $\overline{\mathbb{T}}_n$. The operator $\overline{T}_n \colon \mathcal{L}(\mathcal{X}) \to \mathcal{L}(\mathcal{X})$ is defined by

$$\overline{T}_n h(x_n) := \overline{E}_n(h|x_n) \quad (10)$$

for all real-valued maps $h$ on $\mathcal{X}$, and all $x_n$ in $\mathcal{X}$. In other words, $\overline{T}_n h$ is the real-valued map on $\mathcal{X}$, whose value $\overline{T}_n h(x_n)$ in $x_n \in \mathcal{X}$ is the conditional upper expectation of the random variable $h(X(n+1))$, given that the system is in state $x_n$ at time $n$. More generally, we also consider the maps $\overline{\mathbb{T}}_n$ from $\mathcal{L}(\mathcal{X}^{n+1})$ to $\mathcal{L}(\mathcal{X}^n)$, defined by

$$\overline{\mathbb{T}}_n f(x_{1:n}) := \overline{T}_n f(x_{1:n}, \cdot)(x_n) = \overline{E}_n(f(x_{1:n}, \cdot)|x_n) \quad (11)$$

for all $x_{1:n}$ in $\mathcal{X}^n$ and all real-valued maps $f$ on $\mathcal{X}^{n+1}$. Of course, we can also consider lower expectations and lower transition operators, which are related to the upper expectations and upper transition operators by conjugacy.

The upper expectations $\overline{E}(\cdot|x_{1:n})$ and $\overline{E}$ on $\mathcal{L}(\mathcal{X}^N)$ can be calculated very easily by backwards recursion.

**Theorem 2** (Concatenation Formula). *For any $x_{1:n}$ in $\mathcal{X}^n$, $n = 1, \ldots, N-1$, and for any real-valued map $f$ on $\mathcal{X}^N$:*

$$\overline{E}(f|x_{1:n}) = \overline{\mathbb{T}}_n \overline{\mathbb{T}}_{n+1} \ldots \overline{\mathbb{T}}_{N-1} f(x_{1:n}) \quad (12)$$

$$\overline{E}(f) = \overline{E}_1(\overline{\mathbb{T}}_1 \overline{\mathbb{T}}_2 \ldots \overline{\mathbb{T}}_{N-1} f). \quad (13)$$

Call, for any non-empty subset $I$ of $\{1\ldots,N\}$, a real-valued map $f$ on $\mathscr{X}^N$ $I$-*measurable* if $f(x_{1:N}) = f(z_{1:N})$ for all $x_{1:N}$ and $z_{1:N}$ in $\mathscr{X}^N$ such that $x_k = z_k$ for all $k \in I$. In other words, an $I$-measurable $f$ only depends on the states $X(k)$ at times $k \in I$. As an example, an $\{n\}$-measurable map $h$ only depends on the state $X(n)$ at time $n$, and we identify it with a map on $\mathscr{X}$ (but remember that it acts on states at time $n$). The following proposition tells us that all upper conditional expectations satisfy a Markov condition.

**Proposition 3** (Markov Condition). *Consider an imprecise Markov chain with finite state set $\mathscr{X}$ and time horizon $N$. Fix $n \in \{1,\ldots,N-1\}$. Let $x_{1:n-1}$ and $z_{1:n-1}$ be arbitrary elements of $\mathscr{X}^{n-1}$, and let $x_n \in \mathscr{X}$. Let $f$ be any $\{n, n+1, \ldots, N\}$-measurable real-valued map on $\mathscr{X}^N$. Then it holds that $\overline{E}(f|x_{1:n-1}, x_n) = \overline{E}(f|z_{1:n-1}, x_n)$, and therefore we may write $\overline{E}(f|x_{1:n-1}, x_n) = \overline{E}_{|n}(f|x_n)$.*

The index '$|n$' is meant to make clear that we are considering an expectation conditional on $X(n) = x_n$.

If we apply the joint upper expectation $\overline{E}$ to maps $h$ that only depend on the state $X(n)$ at time $n$, we get the *marginal upper expectation* $\overline{E}_n(h) := \overline{E}(h)$, which is a model for the uncertainty about the state $X(n)$ at time $n$. More generally, taking into account Proposition 3, we use the notation $\overline{E}_{n|\ell}(h|x_\ell) := \overline{E}_{|\ell}(h|x_\ell)$ for the upper expectation of $h(X(n))$, conditional on $X(\ell) = x_\ell$ with $1 \leq \ell < n$. With notations established in Eq. (7), $\overline{E}_{n+1|n}(h|x_n) = \overline{E}_n(h|x_n) = \overline{T}_n h(x_n)$. Such expectations can be found using simpler recursion formulae than Eqs. (12) and (13), as they are based on the simpler upper transition operators $\overline{T}_k$.

**Corollary 4.** *For any real-valued map $h$ on $\mathscr{X}$, and for any $1 \leq \ell < n \leq N$ and all $x_\ell$ in $\mathscr{X}$:*

$$\overline{E}_{n|\ell}(h|x_\ell) = \overline{T}_\ell \overline{T}_{\ell+1} \ldots \overline{T}_{n-1} h(x_\ell),$$
$$\overline{E}_n(h) = \overline{E}_1(\overline{T}_1 \overline{T}_2 \ldots \overline{T}_{n-1} h). \quad (14)$$

This offers a reason for formulating our theory in terms of real-valued maps rather than events: suppose we want to calculate the upper probability $\overline{E}_n(A)$ that the state $X(n)$ at time $n$ belongs to the set $A$. According to Eq. (14), $\overline{E}_n(A) = \overline{E}_1(\overline{T}_1 \ldots \overline{T}_{n-1} I_A)$, and even if $\overline{T}_{n-1}$ can still be calculated using upper probabilities only, it will generally assume values other than 0 and 1, and therefore will not be the indicator of some event. Already after one step, i.e., in order to calculate $\overline{T}_{n-2}\overline{T}_{n-1} I_A$, we need to leave the ambit of events, and turn to the more general real-valued maps; even if we only want to calculate upper *probabilities* after $n$ steps. But for joint upper and lower probability mass functions, however, we can remain within the ambit of events:

**Proposition 5** (Chapman–Kolmogorov Equations). *For an imprecise Markov chain, we have for all $1 \leq n < m \leq N$ and all $(x_n, x_{n+1:m}) \in \mathscr{X}^{m-n+1}$ that*

$$\overline{E}_{|n}(\{x_{n+1:m}\}|x_n) = \prod_{k=n}^{m-1} \overline{T}_k I_{\{x_{k+1}\}}(x_k), \quad (15)$$

*and for all $1 \leq m \leq N$ and all $x_{1:m} \in \mathscr{X}^m$ that*

$$\overline{E}(\{x_{1:m}\}) = \overline{E}_1(\{x_1\}) \prod_{k=1}^{m-1} \overline{T}_k I_{\{x_{k+1}\}}(x_k). \quad (16)$$

*Analogous expressions hold for the lower expectations.*

## 4 Convergence for imprecise Markov chains

Let us now consider a *stationary* imprecise Markov chain with infinite horizon, meaning that $\overline{T}_1 = \overline{T}_2 = \cdots = \overline{T}_n = \ldots =: \overline{T}$. Analogous to the precise case, we define the regularity condition for the upper transition operator $\overline{T}$. It will turn out to be a sufficient condition for convergence.

**Definition 6** (Regularity for upper transition operators). *We call an upper transition operator $\overline{T}$ regular if there is some $n \in \mathbb{N}$ such that $\min \overline{T}^n I_{\{y\}} > 0$ for all $y$ in $\mathscr{X}$.*

We call an upper expectation $\overline{E}$ on $\mathscr{L}(\mathscr{X})$ $\overline{T}$-*invariant* whenever $\overline{E} \circ \overline{T} = \overline{E}$, i.e., if $\overline{E}(\overline{T}h) = \overline{E}(h)$ for all $h \in \mathscr{L}(\mathscr{X})$. With this definition, we can formulate the upper expectation form of the Perron–Frobenius theorem.

**Theorem 7** (Perron–Frobenius Theorem, Upper Expectation Form). *Consider a stationary imprecise Markov chain with finite state set $\mathscr{X}$ and an upper transition operator $\overline{T}$ that is regular. Then for every initial upper expectation $\overline{E}_1$, the upper expectation $\overline{E}_n = \overline{E}_1 \circ \overline{T}^{n-1}$ for the state at time $n$ converges point-wise to the same upper expectation $\overline{E}_\infty$:*

$$\lim_{n \to \infty} \overline{E}_n(h) = \lim_{n \to \infty} \overline{E}_1(\overline{T}^{n-1}h) =: \overline{E}_\infty(h)$$

*for all $h$ in $\mathscr{L}(\mathscr{X})$. Moreover, the limit upper expectation $\overline{E}_\infty$ is the only $\overline{T}$-invariant upper expectation on $\mathscr{L}(\mathscr{X})$.*

The classical Perron–Frobenius Theorem (Theorem 1) is of course a special case of our Theorem 7. Škulj (2007) uses a different, credal set approach to prove a similar (but much weaker) result for imprecise Markov chains with regular *lower* transition operators $\underline{T}$. He also proves a convergence result for conservative (too large) approximations of the $\overline{E}_n$, in the special case that $\overline{T}$ is regular but 2-alternating; see Section 5.3 for further details.

## 5 Examples

In this section, we indicate how the above theory can be applied in a number of practical situations, where the upper expectations are of some special types, described in the literature on imprecise probabilities. We present concrete and explicit examples, as well as a number of simulations.

### 5.1 Contamination models

Suppose we consider a precise stationary Markov chain, with transition operator T. We contaminate it with a vacuous model, i.e., we take a convex mixture with the upper transition operator $I_{\mathscr{X}}$ max. This leads to the upper transition operator $\overline{\mathrm{T}}$, defined by

$$\overline{\mathrm{T}}h = (1-\varepsilon)\mathrm{T}h + I_{\mathscr{X}}\varepsilon\max h, \qquad (17)$$

for all $h \in \mathscr{L}(\mathscr{X})$, where $\varepsilon$ is some constant in the open real interval $(0,1)$. The underlying idea is that we consider a specific convex neighbourhood of T. Since for all $x$ in $\mathscr{X}$, $\min \overline{\mathrm{T}}I_{\{x\}} = (1-\varepsilon)\min \mathrm{T}I_{\{x\}} + \varepsilon > 0$, this upper transition operator is always regular, regardless of whether T is! We infer from Theorem 7 that, whatever the initial upper expectation operator $\overline{E}_1$ is, the upper expectation operator $\overline{E}_n$ for the state $X(n)$ will always converge to the same $\overline{E}_\infty$.

What is this $\overline{E}_\infty$ is for given T and $\varepsilon$? For any $n \geq 1$, $\overline{\mathrm{T}}^n h = (1-\varepsilon)^n \mathrm{T}^n h + I_{\mathscr{X}} \varepsilon \sum_{k=0}^{n-1}(1-\varepsilon)^k \max \mathrm{T}^k h$, and therefore

$$\overline{E}_{n+1}(h) = (1-\varepsilon)^n \overline{E}_1(\mathrm{T}^n h) + \varepsilon \sum_{k=0}^{n-1}(1-\varepsilon)^k \max \mathrm{T}^k h. \quad (18)$$

If we now let $n \to \infty$, we see that the limit is indeed independent of the initial upper expectation $\overline{E}_1$:

$$\overline{E}_\infty(h) = \varepsilon \sum_{k=0}^{\infty}(1-\varepsilon)^k \max \mathrm{T}^k h. \qquad (19)$$

**Example 5.1** (Contaminating a cycle)**.** Consider for instance $\mathscr{X} = \{a,b\}$, and let the precise Markov chain be the cycle with period 2, with transition operator T given by $\mathrm{T}h(a) = h(b)$ and $\mathrm{T}h(b) = h(a)$. Then $\mathrm{T}^{2n}h = h$ and $\mathrm{T}^{2n+1}h = \mathrm{T}h$, and therefore $\max \mathrm{T}^{2n}h = \max \mathrm{T}^{2n+1}h = \max h$, whence $\overline{E}_\infty(h) = \max h$. ♦

**Example 5.2** (Contaminating a random walk)**.** Consider a random walk, where $\mathscr{X} = \{a,b\}$ and $\mathrm{T}h = I_{\mathscr{X}}\frac{h(a)+h(b)}{2}$. Then we find that

$$\overline{E}_\infty(h) = \varepsilon \max h + (1-\varepsilon)\frac{h(a)+h(b)}{2}. \qquad ♦$$

**Example 5.3** (Another contamination model)**.** To illustrate the convergence properties of an imprecise Markov chain, let us look at a simple numerical example. Again consider $\mathscr{X} = \{a,b\}$ and let the stationary imprecise Markov chain be defined by an initial credal set $\mathscr{M}_1 = \{m \in \Sigma_{\{a,b\}}: 0.6 \leq m(a) \leq 0.9\}$, and a contamination model of the type (17), with $\varepsilon = 0.1$, and for which the precise transition operator T is defined by the Markov matrix $T := \begin{bmatrix} q(a|a) & q(b|a) \\ q(a|b) & q(b|b) \end{bmatrix} = \begin{bmatrix} 0.15 & 0.85 \\ 0.85 & 0.15 \end{bmatrix}$. In Figure 4 we plot the evolution of $\overline{E}_n(\{a\})$ and $\underline{E}_n(\{a\})$, the upper and lower probability for finding the system in state $a$ at time $n$, which can be calculated efficiently using Eq. (18).

For comparison, we also plot the evolution of $E_n(\{a\})$, the probability for finding the system in state $a$ at time $n$, for a (precise) Markov chain defined by probability mass functions that lie on the boundaries of the credal sets defining the above imprecise Markov chain; to wit, its initial mass function is given by $M_1 := \begin{bmatrix} m_1(a) & m_1(b) \end{bmatrix} = \begin{bmatrix} 0.9 & 0.1 \end{bmatrix}$ and its Markov matrix is $\begin{bmatrix} 0.135 & 0.865 \\ 0.865 & 0.135 \end{bmatrix}$. Here $E_\infty(\{a\}) = E_\infty(\{b\}) = 0.5$. ♦

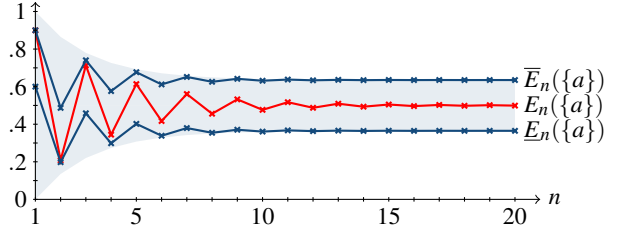

Figure 4: The time evolution of (i) the upper and lower probability of finding the imprecise Markov chain of Example 5.3 in the state $a$ (outer plot marks and connecting lines); and of (ii) the probability of finding the classical Markov chain of Example 5.3 in the state $a$ (inner plot marks and connecting lines). The filled area denotes the hull of the evolution of this probability, under the contamination model of Example 5.3, for all possible initial mass functions.

### 5.2 Belief function models

The contamination models we have just described are a special case of a more general and quite interesting class of models, based on Shafer's notion of a belief function Shafer (1976). We can consider a number of subsets $F_j$, $j = 1, \ldots, n$ of $\mathscr{X}$, and a convex mixture of the vacuous upper expectations relative to these subsets:

$$\overline{E}(h) = \sum_{j=1}^{n} m(F_j) \max_{x \in F_j} h(x), \qquad (20)$$

with $m(F_j) \geq 0$ and $\sum_{j=1}^{n} m(F_j) = 1$. In Shafer's terminology, the sets $F_j$ are called *focal elements*, and the $m(F_j)$'s the *basic probability assignment*.

We can now consider imprecise Markov chains where the local models, attached to the non-terminal situations in the tree, are of this type. The general backwards recursion formulae we have given in Section 3 can then be used in combination with the simple formulae of the type (20) for an efficient calculation of all conditional and joint upper and lower expectations in the tree. We leave this implicit however, and move on to another example, which is rather more popular in the literature.

### 5.3 Models with lower and upper mass functions

An intuitive way to introduce imprecise Markov chains (Kozine and Utkin, 2002; Campos et al., 2003; Škulj, 2006) goes by way of so-called *probability intervals*, studied in a paper by De Campos et al. (1994); see also Walley (1991, Section 4.6.1). It consists in specifying lower and upper bounds for mass functions. Let us explain how this is done in the specific context of Markov chains.

For the initial mass function $m_1$, we specify a lower bound $\underline{m}_1 \colon \mathscr{X} \to \mathbb{R}$, also called a *lower mass function*, and an upper bound $\overline{m}_1 \colon \mathscr{X} \to \mathbb{R}$, called an *upper mass function*.

The credal set $\mathscr{M}_1$ attached to the initial situation, which corresponds to these bounds, is then given by

$$\mathscr{M}_1 := \{m \in \Sigma_{\mathscr{X}} : (\forall x \in \mathscr{X})(\underline{m}_1(x) \leq m(x) \leq \overline{m}_1(x))\}.$$

Similarly, in each non-terminal situation $x_{1:k} \in \mathscr{X}^k$, $k = 1, \ldots, N-1$ we have a credal set $\mathscr{Q}_k(\cdot|x_k)$ that is defined in terms of conditional lower and upper mass functions $\underline{q}_k(\cdot|x_k)$ and $\overline{q}_k(\cdot|x_k)$. Here, for instance, $\underline{q}_k(x_{k+1}|x_k)$ gives a lower bound on the transition probability $q_k(x_{k+1}|x_k)$ to go from state $X(k) = x_k$ to state $X(k+1) = x_{k+1}$ at time $k$.

Under some consistency conditions—see (De Campos et al., 1994) for more details—the upper expectation associated with $\mathscr{M}_1$ is then given in all subsets $A$ of $\mathscr{X}$ by

$$\overline{E}_1(A) = \min\left\{\sum_{z \in A} \overline{m}_1(z), 1 - \sum_{z \in \mathscr{X} \setminus A} \underline{m}_1(z)\right\},$$

This $\overline{E}_1$ is 2-alternating: $\overline{E}_1(A \cup B) + \overline{E}_1(A \cap B) \leq \overline{E}_1(A) + \overline{E}_1(B)$ for all subsets $A$ and $B$ of $\mathscr{X}$. This implies (Walley, 1991, Section 3.2.4) that for all $h \in \mathscr{L}(\mathscr{X})$ the upper expectation $\overline{E}_1(h)$ can be found by Choquet integration:

$$\overline{E}_1(h) = \min h + \int_{\min h}^{\max h} \overline{E}_1(\{z \in \mathscr{X} : h(z) \geq \alpha\}) \, d\alpha, \quad (21)$$

where the integral is a Riemann integral. Similar considerations for the 2-alternating $\overline{E}_k(\cdot|x_k)$ lead to formulae for the upper transition operators $\overline{T}_k$: for all $x_k$ in $\mathscr{X}$,

$$\overline{T}_k I_A(x_k) = \min\left\{\sum_{z \in A} \overline{q}_k(z|x_k), 1 - \sum_{z \in \mathscr{X} \setminus A} \underline{q}_k(z|x_k)\right\} \quad (22)$$

$$\overline{T}_k h(x_k) = \min h + \int_{\min h}^{\max h} \overline{T}_k I_{\{z \in \mathscr{X} : h(z) \geq \alpha\}}(x_k) \, d\alpha. \quad (23)$$

**Example 5.4** (Close to a cycle). Consider a three-state stationary imprecise Markov model with $\mathscr{X} = \{a, b, c\}$ and with marginal and transition probabilities given by probability intervals. It follows from Eqs. (22) and (23) that the upper transition operator $\overline{T}$ is fully determined by the upper and lower Markov matrices:

$$\begin{bmatrix} \underline{q}(a|a) & \underline{q}(b|a) & \underline{q}(c|a) \\ \underline{q}(a|b) & \underline{q}(b|b) & \underline{q}(c|b) \\ \underline{q}(a|c) & \underline{q}(b|c) & \underline{q}(c|c) \end{bmatrix} = \frac{1}{200} \begin{bmatrix} 9 & 9 & 162 \\ 144 & 18 & 18 \\ 9 & 162 & 9 \end{bmatrix}$$

$$\begin{bmatrix} \overline{q}(a|a) & \overline{q}(b|a) & \overline{q}(c|a) \\ \overline{q}(a|b) & \overline{q}(b|b) & \overline{q}(c|b) \\ \overline{q}(a|c) & \overline{q}(b|c) & \overline{q}(c|c) \end{bmatrix} = \frac{1}{200} \begin{bmatrix} 19 & 19 & 172 \\ 154 & 28 & 28 \\ 19 & 172 & 19 \end{bmatrix},$$

where the numerical values are particular to this example. Similarly, the initial upper expectation $\overline{E}_1$ is completely determined by the matrices $\overline{M}_1$ and $\underline{M}_1$:

$$\underline{M}_1 := \begin{bmatrix} \underline{m}_1(a) & \underline{m}_1(b) & \underline{m}_1(c) \end{bmatrix}$$
$$\overline{M}_1 := \begin{bmatrix} \overline{m}_1(a) & \overline{m}_1(b) & \overline{m}_1(c) \end{bmatrix}.$$

In Figure 5, we plot conservative approximations for the credal sets $\mathscr{M}_n$ corresponding to the upper expectation operators $\overline{E}_n$. Each approximation is based on the constraints that can be found by calculating $\underline{E}_1(\overline{T}^{n-1} I_{\{x\}})$ and $\overline{E}_1(\overline{T}^{n-1} I_{\{x\}})$ using the backwards recursion method, for $x = a, b, c$. The $\mathscr{M}_n$ evolve clockwise through the simplex, which is not all that surprising as the lower and upper Markov matrices are quite 'close' to the precise *cyclic* Markov matrix

$$\begin{bmatrix} q(a|a) & q(b|a) & q(c|a) \\ q(a|b) & q(b|b) & q(c|b) \\ q(a|c) & q(b|c) & q(c|c) \end{bmatrix} = \begin{bmatrix} 0 & 0 & 1 \\ 1 & 0 & 0 \\ 0 & 1 & 0 \end{bmatrix}.$$

After a while, the $\mathscr{M}_n$ converge to a limit that is independent of the initial credal set $\mathscr{M}_1$, as can be predicted from the regularity of the upper transition operator. ♦

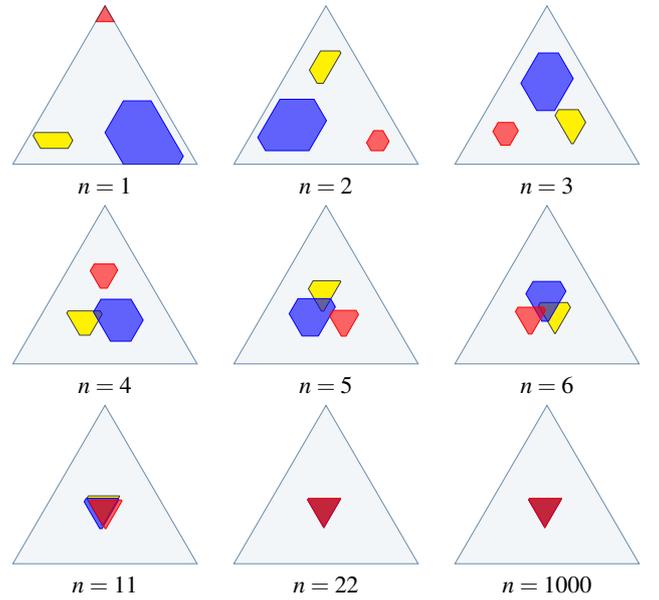

Figure 5: Evolution in the simplex $\Sigma_{\{a,b,c\}}$ of the credal sets $\mathscr{M}_n$ for the near-cyclic transition operator from Example 5.4 for three different choices of the initial credal set $\mathscr{M}_1$.

## 6 Conclusions

Regularity seems to be a reasonably weak condition on the upper transition operator $\overline{T}$ for a stationary imprecise Markov chain, but we have seen that it is strong enough to guarantee that the upper expectation for the state at time $n$ converges to a uniquely $\overline{T}$-invariant upper expectation $\overline{E}_\infty$, regardless of the initial upper expectation $\overline{E}_1$.

Even when the regularity condition is not satisfied, it is not so hard to see that any upper transition operator $\overline{T}$ is still *non-expansive* under the supremum norm given for every $h \in \mathscr{L}(\mathscr{X})$ by $\|h\|_\infty := \max |h|$:

$$\|\overline{T}g - \overline{T}h\|_\infty \leq \|g - h\|_\infty,$$

Moreover, the sequence $\|\overline{T}^n h\|_\infty$ is bounded because $\|\overline{T}^n h\|_\infty \leq \|h\|_\infty$. It then follows from non-linear Perron–Frobenius theory (Sine, 1990; Nussbaum et al., 1998) that the sequence $\overline{T}^n h$ has a periodic limit cycle. More precisely, there is a $\xi_h \in \mathscr{L}(\mathscr{X})$ such that $\overline{T}^{p_h} \xi_h = \xi_h$ i.e., $\xi_h$ is a *periodic point* of $\overline{T}$ with (smallest) *period* $p_h$, and such that $\overline{T}^{np_h} h \to \xi_h$ (point-wise) as $n \to \infty$. It would be a very interesting topic for further research to study the nature of the periods and periodic points of upper transition operators.

In our discussions, for instance in Section 3, we have consistently used the sensitivity analysis interpretation of imprecise probability models such as upper expectations. Upper and lower expectations can also be given another, so-called *behavioural* interpretation, in terms of some subjects dispositions towards accepting risky transactions. This is for instance Walley's (1991) preferred approach. The results we have derived here remain valid on that alternative interpretation, and the concatenation formulae (12) and (13) can then be shown to be special cases of so-called *marginal extension* procedure (Miranda and De Cooman, 2007), which provides the most conservative coherent (i.e., rational) inferences from the local predictive models $\overline{T}_k$ to general lower and upper expectations. In another paper (De Cooman and Hermans, 2008), we give more details about how to approach a process theory using imprecise probabilities on a behavioural interpretation.